\documentclass[10pt,twocolumn,letterpaper]{article}

\usepackage{tex/cvpr}
\usepackage{times}
\usepackage{epsfig}
\usepackage{graphicx}
\usepackage{amsmath}
\usepackage{amssymb}

\usepackage[pagebackref=true,breaklinks=true,letterpaper=true,colorlinks,bookmarks=false]{hyperref}

\usepackage{times}
\usepackage{helvet}
\usepackage{courier}
\usepackage{graphicx}
\usepackage{bm}
\usepackage{amsmath,amssymb} 
\usepackage{color}
\usepackage{bbm}
\usepackage{epstopdf}
\usepackage{caption}
\usepackage{subcaption}
\usepackage{enumitem}
\usepackage{calc}
\usepackage{multirow}
\usepackage{xspace}
\usepackage{booktabs}
\usepackage{mathrsfs}
\usepackage{array}
\usepackage[font=small,skip=4pt]{caption}

\newcommand{\tabref}[1]{Tab\onedot~\ref{#1}}

\newcommand{\ve}[1]{{\mathbf #1}} 
\newcommand{\hua}[1]{{\mathcal #1}}
\newcommand{\scr}[1]{{\mathcal #1}}

\newcommand{\thickhline}{%
    \noalign {\ifnum 0=`}\fi \hrule height 1pt
    \futurelet \reserved@a \@xhline
}

\DeclareRobustCommand\onedot{\futurelet\@let@token\@onedot}
\def\onedot{\ifx\@let@token.\else.\null\fi\xspace}
\def\eg{\emph{e.g.}}

\def\etc{\emph{etc}\onedot}

\def\wrt{w.r.t\onedot}

\def\etal{\emph{et al.}}
\cvprfinalcopy 
\def\cvprPaperID{1681} 

\ifcvprfinal\fi
\begin{document}
\title{Activity Driven Weakly Supervised Object Detection}

\author{
Zhenheng Yang$^{1}$~~Dhruv Mahajan$^{2}$~~Deepti Ghadiyaram$^{2}$~~Ram Nevatia$^{1}$\\
~~Vignesh Ramanathan$^{2}$\\
\\
$^{1}$University of Southern California~~$^{2}$Facebook AI~~\\
}

\maketitle
\vspace{-0.5\baselineskip}
\begin{abstract}
\vspace{-0.6\baselineskip}
    Weakly supervised object detection aims at reducing the amount of supervision required to train detection models. Such models are traditionally learned from images/videos labelled only with the object class and not the object bounding box. In our work, we try to leverage not only the object class labels but also the action labels associated with the data. We show that the action depicted in the image/video can provide strong cues about the location of the associated object. We learn a spatial prior for the object dependent on the action (\eg~``ball" is closer to ``leg of the person" in ``kicking ball"), and incorporate this prior to simultaneously train a joint object detection and action classification model. We conducted experiments on both video datasets and image datasets to evaluate the performance of our weakly supervised object detection model. Our approach outperformed the current state-of-the-art (SOTA) method by more than 6\% in mAP on the Charades video dataset.


    
\vspace{-0.3\baselineskip}
\end{abstract}

\vspace{-1.0\baselineskip}
\section{Introduction}
\vspace{-0.4\baselineskip}
\label{sec:intro}

Deep learning techniques and development of large datasets have been vital to the success of image and video classification models. One of the main challenges in extending this success to object detection is the difficulty in collecting fully labelled object detection datasets. Unlike classification labels, detection labels (object bounding boxes) are more tedious to annotate. This is even more challenging in the video domain due to the added complexity of annotating along the temporal dimension.

\begin{figure}
\vspace{-0.5\baselineskip}
\includegraphics[width=0.48\textwidth]{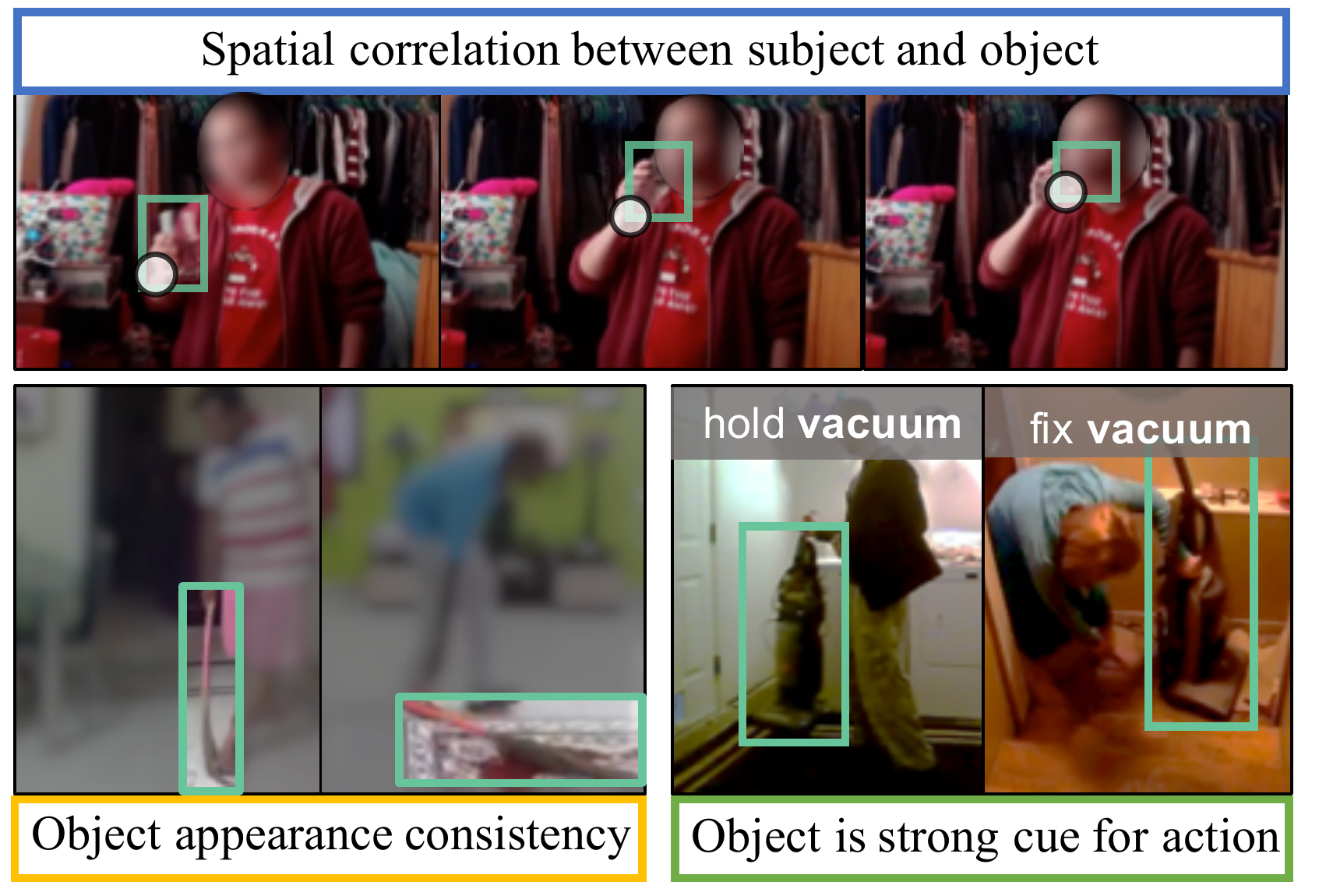}
\caption{Our framework is built upon three observations we draw: (1) there is spatial dependence between the subject and the interacted object; (2) the object appearance is consistent across different training samples and across different actions involving the same object; (3) the most informative object about the action is the one mentioned in the action.}
\label{fig:poor_figure}
\vspace{-1.2\baselineskip}
\end{figure}

\begin{figure}
\vspace{-0.4\baselineskip}
\centering
\includegraphics[width=0.45\textwidth]{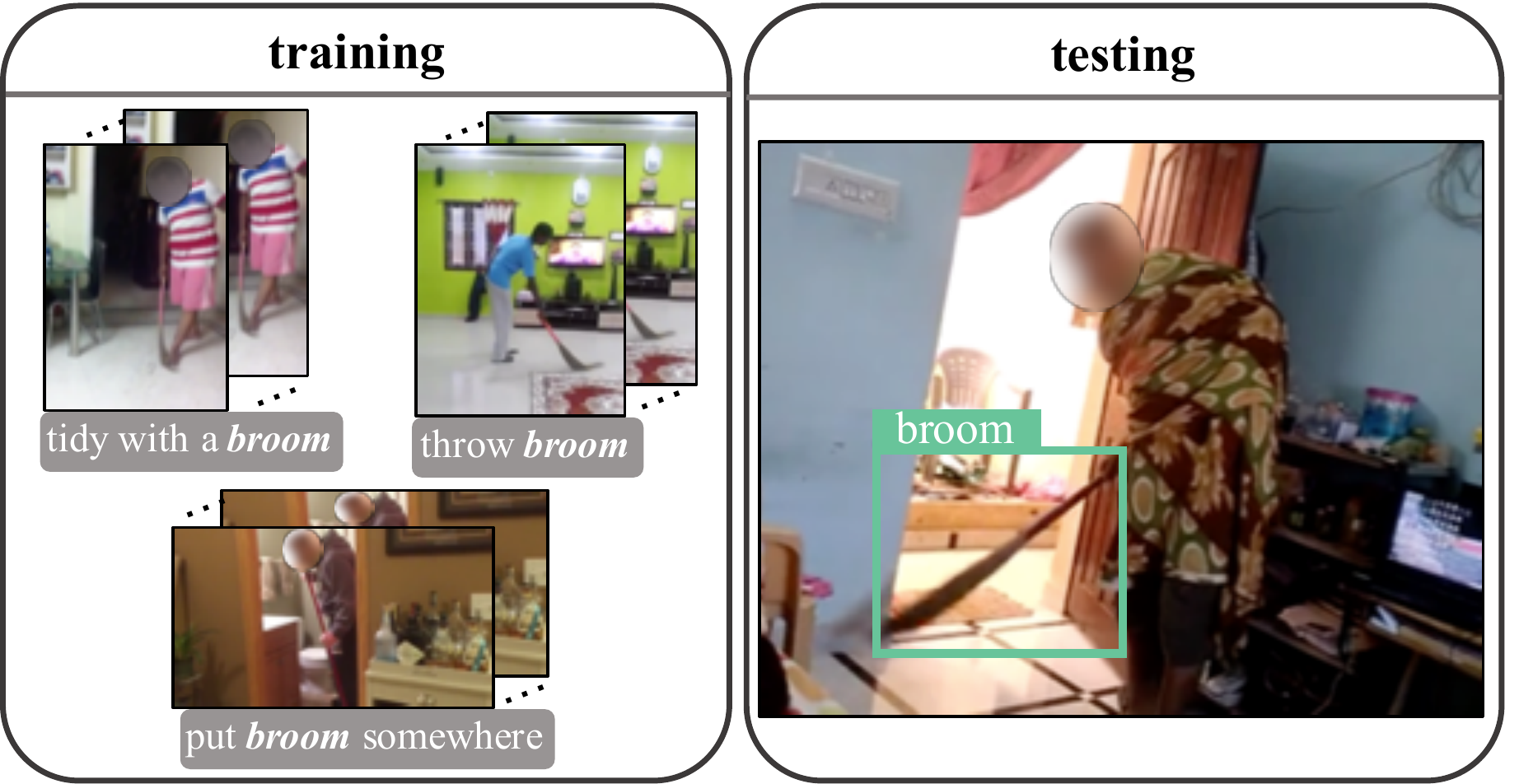}
\caption{Setting of the action-driven weakly supervised object detection task. Training samples include videos or images with action class labels (left). The inference is conducted on single frame/image for object localization and classification (right).}
\label{fig:task_definition}
\vspace{-1.0\baselineskip}
\end{figure}

On the other hand, there are a large number of video and image datasets \cite{sigurdsson2016hollywood,gu2017ava,carreira2017quo,chao2018learning,Damen2018EPICKITCHENS} labelled with human actions which are centered around objects. Action labels provide strong cues about the location of the corresponding objects in a scene (Fig.~\ref{fig:poor_figure}) and could act as weak supervision for object detection. In light of this, we investigate the idea of learning object detectors from data labelled only with action classes as shown in Fig. \ref{fig:task_definition}.

All images/videos associated with an action contain the object mentioned in the action (\eg ``cup'' in the action ``drink from cup''). Yuan et al. \cite{yuan2017temporal} leveraged this property to learn object detection from videos of corresponding actions. However, the actions (``drink from" in above example) themselves are not utilized in this work. On the other hand, the spatial location, appearance and movement of objects in a scene are dependent on the action performed with the object. The key contribution of our work is to leverage this intuition to build better object detection models.

Specifically, we have three observations (see Fig. \ref{fig:poor_figure}): (1) There is spatial dependence between the position of a person and the object mentioned in the action, \eg in action ``hold cup'', the location of \textit{cup} is tightly correlated with the location of the \textit{hand}. This could provide a strong prior for the object; (2) The object appearance is consistent across images and videos of action classes which involve the object; (3) Detecting the object should help in predicting the action and vice-versa.


The above observations can be used to address one of the main challenges of weakly supervised detection: the presence of a large search space for object bounding boxes during training. Each training image/video has many candidate object bounding boxes (object proposals). In our weakly supervised setting, the only label we have is that, one of these candidates should correspond to the object mentioned in the action. The training algorithm is required to automatically identify the correct object bounding box from this large set of candidates. In our approach, we narrow down this search by incorporating the three observations in our model. In particular, we (1) explicitly learn the spatial prior of objects \wrt~the human in different actions; (2) train a generic object classifier for modeling object appearances across different actions; (3) jointly learn the action classifier and associated object classifier.



We conducted comprehensive experiments over two video datasets: Charades \cite{sigurdsson2016hollywood}, EPIC KITCHENS \cite{Damen2018EPICKITCHENS} and an image dataset: HICO-DET \cite{chao2018learning}. Our method outperforms the previous methods \cite{bilen2016weakly,yuan2017temporal,tang2018pcl} by a large margin on all datasets. Specifically, we have achieved a ~6\% mAP boost on Charades compared to current state-of-the-art weakly supervised models for videos. Visualization results and ablation experiments show the effectiveness of each module in our approach.

\vspace{-0.5\baselineskip}
\section{Related Work}
\vspace{-0.5\baselineskip}
\label{sec:related}
In this section, we briefly overview some related research topics and how we are motivated by these works.

\textbf{~Supervised Object detection.}
Object detection is a very active research topic in the computer vision field. There has been significant progress in the recent years with the advances of deep learning. R-CNN \cite{girshick14CVPR} is the first work that introduces CNN features to object detection. 
A sequence of later works are developed based on R-CNN. Fast R-CNN \cite{girshick2015fast} accelerates R-CNN by introducing an ROI pooling layer and improve the performance by applying proposal classification and bounding box regression jointly. Faster R-CNN \cite{ren2015faster} further improves the speed and accuracy by replacing the proposal generation stage with a learnable network: region proposal network and the whole framework is trained in an end-to-end fashion. Mask R-CNN \cite{he2017mask} proposed to add a segmentation branch and achieved the state-of-the-art (SoTA) performance. All the methods require full object bounding box annotations and mask R-CNN requires dense segmentation labels.

\textbf{Weakly supervised object detection.}
The fully supervised object detection methods rely heavily on large scale bounding box annotations, which is inefficient and labor consuming. To alleviate this issue, there have been various weakly-supervised works \cite{cinbis2017weakly,song2014learning,bilen2016weakly,kantorov2016contextlocnet,jie2017deep,peyre2017weakly,shi2017weakly,wei2018ts2c,oh2017exploiting,bai2017multiple,singh2017hide,zhu2017soft,zhang2018zigzag,srikantha2017weak,zhou2018weakly,zhang2018w2f,diba2017weakly,shen2018weakly,wan2018min,zhang2018adversarial} that leverage the more efficient image-level object class annotations. Weakly supervised deep detection networks (WSDDN) \cite{bilen2016weakly} proposed an end-to-end architecture to perform region selection and classification simultaneously. It is achieved by separately performing classification and detection headers and the supervision comes from a combination classification score. ContextLocNet \cite{kantorov2016contextlocnet} further improves WSDDN by taking contextual region into consideration. Beyond the image domain, another line of research works \cite{kwak2015unsupervised,wang2015unsupervised} try to leverage the temporal information in videos to facilitate the weakly supervised object detection. Kwak \etal~\cite{kwak2015unsupervised}  proposed to discover the object appearance presentation across videos and then track the object in temporal space for supervision. Wang \etal~\cite{wang2015unsupervised} perform unsupervised tracking on videos and then cluster similar deep features to form visual representation.

Yuan \etal~\cite{yuan2017temporal} proposed a much more efficient action-driven weakly supervised object detection setting which aims to learn the object appearance representation given only videos with clip-level action class labels. They proposed to first extract spatial features from object proposals. The features are then updated using long short-term memory (LSTM) \cite{hochreiter1997long} applied on neighboring frames. The frame-level object classification loss is computed on the updated features. We implemented the same setting as in \cite{yuan2017temporal}: pipeline trained on videos/images with only action labels and test on images. Unlike TD-LSTM \cite{yuan2017temporal} that only leverages object class information, we propose to jointly exploit both action and object class labels. Considering all actions are interactions between person and objects, we incorporate human pose estimation into the framework.

\textbf{Activity recognition}
There are a variety of works in the field of action recognition \cite{wang2017untrimmednets,maji2011action,gkioxari2015contextual,alayrac2017joint,gao2017turn,gan2016webly,gao2017cascaded}. Maji \etal~\cite{maji2011action} train action specific poselets that are then classified using SVMs. The contextual cues are captured by explicitly detecting objects and exploiting action labels of other people in the image. R*CNN \cite{gkioxari2015contextual} proposed to implicitly model the main objects. The features from both person region and object proposal regions are extracted and a fusion of classification scores from these two types of features is used for action classification loss. R*CNN showed that the most informative object in the scene is the object mentioned in action class. We are inspired by the similar idea to jointly consider the action and object lables.

\begin{figure*}[ht!]
\centering
\vspace{-0.5\baselineskip}
\includegraphics[width=0.95\textwidth]{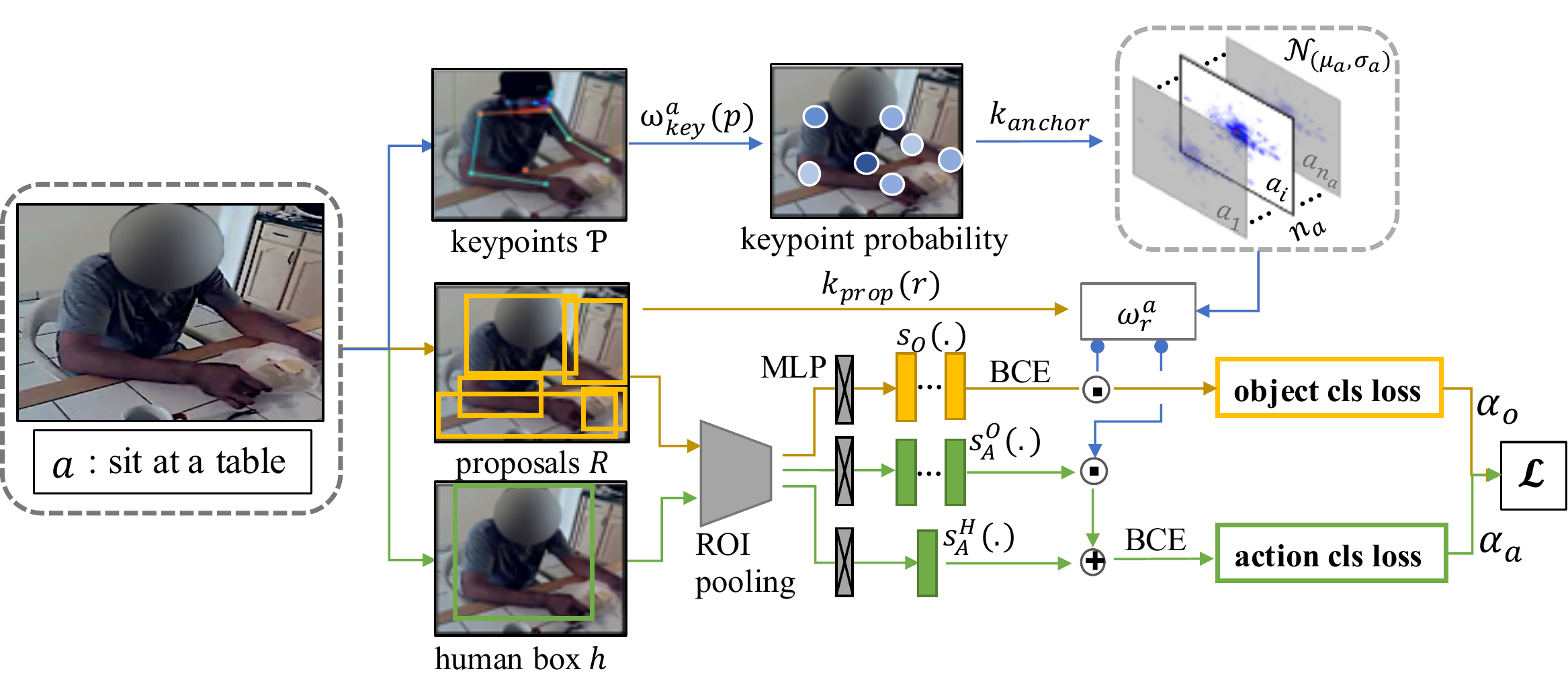}
\caption{The diagram of our framework. There are three streams in the proposed framework: object spatial prior module (colored in blue), object classification stream (colored in yellow) and action classification module (colored in green). We incorporate human keypoint detection into the framework and jointly leverage action and object labels.}
\vspace{-1.2\baselineskip}
\label{fig:pipeline}
\end{figure*}

\textbf{Human-object interaction}
There are mainly two tasks in the human-object interaction (HOI) topic: HOI recoginition and HOI detection. HOI recognition aims at recognizing the interaction between subject and object. There have been a surge of works \cite{delaitre2010recognizing,gupta2009observing,yao2010modeling,shen2018scaling} on HOI recognition since 2009.
HOI detection task aims at localizing subject and object and also recognizing HOI class. Chao \etal~\cite{chao2018learning} proposed a three-stream network for this task, one stream each for person detection, object detection and person-object pair wise classification respectively. Gkioxari \etal~\cite{gkioxari2017detecting} model the interaction with shared weights between human centric branch and interaction branch. Kalogeiton \etal~\cite{kalogeiton2017joint} proposed to jointly learn the object and action (\eg~\textit{dog running}). All works have shown that jointly learning the object, person localization and HOI/action classification benefits the performance.

\vspace{-0.5\baselineskip}
\section{Approach}
\vspace{-0.4\baselineskip}
\label{sec:approach}

The main challenge of weakly supervised detection is the lack of bounding box information during training and the availability of only image/video level labels. This problem is typically handled in a Multiple Instance Learning (MIL) setting \cite{bilen2016weakly,cinbis2017weakly}, where the training method implicitly chooses the best bounding box from a set of candidate proposals in the image/video to explain the overall image/video label.

However, in practice the number of candidate object proposals can be quite large, making the problem challenging. In our work, we address this issue by imposing additional constraints on the choice of the best object bounding box based on the location prior of the object \wrt the human and the importance of the chosen object proposal for action classification. In practice, we model each of these as three different streams in our model which finally contribute to a single action classification loss and an object classification loss. 
Note that in our work we assume that a pre-trained person detection model and human keypoint detection model are available to extract the signals needed for capturing human-object dependence.

\vspace{-0.4\baselineskip}
\subsection{Framework}
\vspace{-0.5\baselineskip}

Formally, for a training sample (video clip or image), the action label $a$ is provided. The action $a$ belongs to a pre-defined set of actions $a \in \scr{A}$, which is of size $n_{a}$: $||\scr{A}||=n_{a}$. We assume that all human actions are interactive and there is one object involved in each action. For example, the object \textit{cup} appears in the action \textit{holding a cup}.
The object class associated with action $a$ is denoted by $o_a$ and there are $n_o$ object classes in total: $o_a \in \scr{O}, ||\scr{O}||=n_o$.

A pre-trained human detector \cite{he2017mask} and pose estimation network \cite{wei2016convolutional} are used to extract human bounding box $h$ and keypoint locations $k(p), p \in \hua{P}$ where $\hua{P}$ represents the set of human keypoints. For training samples with multiple people, we pick the detection result with the highest detection confidence. The object proposals $R$ are extracted. We remove proposals with high overlap ($IoU > \theta_h$) with human region $h$ and we keep the top $n_r$ with highest confidence.

Our model has three streams which are explained in detail in the next sections. An overview of our models is shown in Fig.~
\ref{fig:pipeline}. The first stream models the spatial prior of the object \wrt to human keypoints in each action. The prior is used to construct an object classification stream which weights the object classification losses of different proposals in an image/video. The weights and features from the object proposals, along with features of human bounding box, are used to construct an action classification loss. The combined loss from action classification and object classification is minimized during training.

\vspace{-0.3\baselineskip}
\subsection{Object spatial prior}
\vspace{-0.3\baselineskip}
The object spatial prior is modeled in two stages: (1) given an action class $a$ and keypoint detection results $\hua{P}$, we estimate an anchor location based on a weighted combination of the keypoint locations; (2) given the action class and the anchor position, the position of the object is modeled as a normal distribution \wrt the anchor point. This is based on our observation that for a given action, certain human keypoints provide strong location priors for the object locations(``hand" for drinking from a cup, ``foot" for kicking a ball etc.).

The anchor location $k_{anchor}$ is calculated as a weighted sum of all keypoint locations. The keypoint weight is modeled with a probability vector $w_{key}^a(p),~ p \in \hua{P}$ for the action class $a$.
\vspace{-0.3\baselineskip}
{\small
\begin{align}
\label{eqn:anchor}
k_{anchor} = \sum_{p \in \hua{P}} w_{key}^a(p) k(p)
\end{align}}
\vspace{-0.6\baselineskip}

\noindent where $k(p)$ is the detected position of the keypoint $p$ in the training image/video.
Given the action class $a$, the weight of object location \wrt the anchor location is modeled with a learned normal distribution: $\scr{N}_{(\ve{\mu}_a, \ve{\sigma}_a)}~~\ve{\mu}_a\in \mathbbm{R}^2, \ve{\sigma}_a\in \mathbbm{R}^{(2\times2)}$. $\ve{\mu}_a$ represents the mean location of the object \wrt the anchor and $\ve{\sigma}_a$ represents the variance. This distribution is used to calculate the object location probabilities of different locations. Specifically, the probability of an object being at the location of a proposal $r \in R$ for an action class $a$ is 
\vspace{-0.3\baselineskip}
{\small
\begin{align}
w_r^{a}=\scr{N}_{(\mu_a, \sigma_a)}(k_{prop}(r)- k_{anchor})
\label{eqn:obj_weight}
\end{align}}
\vspace{-0.5\baselineskip}

\noindent where $k_{prop}(r)$ is the center of the proposal $r$. Note that the distributions $w_{key},~~\scr{N}_{(\mu_a, \sigma_a)}$ are learned automatically during training.
\vspace{-0.2\baselineskip}
\subsection{Object classification}
\vspace{-0.2\baselineskip}
For each proposal $r \in R$ in a training sample, we compute an object classification score for each object $o$: $s_O(r;o)$. Here $s_O$ corresponds to an ROI-pooling layer followed by a Multi Layer Perceptron (MLP) which classifies the input region into $n_o$ object classes. Apart from only leveraging image-level object labels for classification \cite{bilen2016weakly,kantorov2016contextlocnet}, the spatial location weights from previous section are also used to guide the selection of the object proposal. Formally, the binary cross-entropy (BCE) loss is calculated on each proposal region, against the image-level object class ground truth. The BCE losses are weighted by the location probabilities of different proposals and the weighted sum is used to compute object classification loss:
\vspace{-0.3\baselineskip}
{\small
\begin{align}
\label{eqn:obj_cls_loss}
&\hua{L}_{obj} = -\frac{1}{n_r}\sum_{r \in R} w_r^{a} \cdot \hua{L}_{o} (r), \nonumber \\
&\hua{L}_{o}(r)\! =\! \frac{1}{n_o} \sum_{o \in O} y_o log(P(o|r))\!+\!(1\!-\!y_o)log(1\!-\!P(o|r)), \nonumber\\
&P(o|r) = \frac{exp(s_O(r;o))}{\sum_{o \in O}exp(s_O(r; o))},
\end{align}}
\vspace{-0.6\baselineskip}

\noindent where $y_o$ is the binary object classification label for the object $o$. Note that $y_o$ is non-zero only for the object mentioned in the action corresponding to the image/video.
\vspace{-0.2\baselineskip}
\subsection{Action classification}
\vspace{-0.3\baselineskip}
For the task of action recognition, especially for interactive actions as in our task, both the person and the object appearances are vital cues. As indicated in \cite{gkioxari2015contextual}, the spatial location of the most informative object can be mined from action recognition task. We incorporate a similar idea into the action classification stream by fusing features from the proposal regions and person region. Formally, for a training instance with action label $a$, the appearance features of both person region $h$ and proposal regions $R$ are extracted, and then classified to $n_a$-dimension action classification scores: $s_A^O(r;a), r \in R$ and $s_A^H(h;a)$. Here $s_A^H,~s_A^O$ correspond to an ROI-pooling layer followed by a Multi Layer Perceptron (MLP). The weights and biases of the MLP are learned during training. The final proposal score is computed as an average of action classficiation scores weighted by the spatial prior probabilities as in the previous section. This ensures that only scores from the most relevant proposals are given a higher weight. The sum of action classification scores from object proposals and person regions is used to compute the final BCE action classification loss. The loss is computed as follows:
\vspace{-0.3\baselineskip}
{\small
\begin{align}
\label{eqn:act_cls_loss}
\vspace{-1.3\baselineskip}
&\hua{L}_{act}\!\!\! =\!\! -\frac{1}{n_a}\!\!\sum_{a \in \scr{A}} y_a log(P(a))\!\!+\!\!(1\!-\!y_a) log(1\!\!-\!\!P(a)), \nonumber\\
&P(a) = \frac{exp\left(s_A^H(h;a)+\sum_{r \in R}  w_r^{a} s_A^O(r;a)\right)}{\sum_{a \in \scr{A}}exp\left(s_A^H(h;a)+\sum_{r\in R}  w_r^{a} s_A^O(r;a)\right)},
\vspace{-1.3\baselineskip}
\end{align}}
\vspace{-0.6\baselineskip}

\noindent where $y_a$ is the binary action classification label for the action $a$.

\vspace{-0.3\baselineskip}
\subsection{Temporal pooling for videos}
\vspace{-0.3\baselineskip}
Our experiments are conducted on both video and image datasets, thus the training samples can be video sequences or static images with action labels. For models trained with video clips, we adopt a few pre-processing steps and also pool scores across the temporal dimension to improve person detection and object proposal quality. Formally, $n$ frames are uniformly sampled from the training clip, followed by person detection and object proposal generation for the sampled frames. The object proposals as well as person bounding boxes across the frames are then connected by an optimization based linking method \cite{gkioxari2015finding,yang2017spatio} to form object proposal tubelets and person tubelets respectively. We observed that temporal linking of proposals avoids spurious proposals and leads to more robust features from the proposals. These are fed as inputs into the object classification and action classification streams. Temporal pooling is used to aggregate classification scores across the person and object tubelets. The pooled scores are finally used for loss computation as before. 
\vspace{-0.4\baselineskip}
\subsection{Loss terms}
\vspace{-0.4\baselineskip}
The combined loss is a weighted sum of both classification loss terms.
\vspace{-0.3\baselineskip}
\begin{align}
\label{eqn:final_loss_term}
\hua{L} = \alpha_o\hua{L}_{obj} + \alpha_a\hua{L}_{act}
\end{align}
The hyper-parameters $\alpha_o$ and $\alpha_a$ are weights to trade off the relative importance of object classification and action classification in the pipeline.
\vspace{-0.5\baselineskip}
\subsection{Inference}
\vspace{-0.5\baselineskip}
During testing, object proposals are firstly extracted on the test sample. The trained object classifier ($s_O$) is applied on each proposal region to obtain the object classification scores ($P(o|r)$). Then the non-maximal suppression (NMS) is applied and the object proposals with higher classification scores than the threshold are preserved as detection results.

\vspace{-0.3\baselineskip}
\section{Experiments}
\vspace{-0.3\baselineskip}
\label{sec:evaluation}

Our method is applicable to both video and image domains. We require only human action label annotations for training. Object bounding box annotations are used only during evaluation. Code will be released in the Github repository \footnote{\url{https://github.com/zhenheny/Activity-Driven-Weakly-Supervised-Object-Detection}}.


 \textbf{Video datasets}: The Charades dataset \cite{sigurdsson2016hollywood} includes 9,848 videos of 157 action classes, among which, 66 are interactive actions with objects. There are on average 6.8 action labels for a video. The official Charades dataset doesn't provide object bounding box annotations and we use the annotations released by \cite{yuan2017temporal}.
In the released annotations, 1,812 test videos are down-sampled to 1 frame per second (fps) and 17 object classes are labeled with bounding boxes on these frames. There are 3.4 bounding box annotations per frame on average. We follow the same practice as in \cite{yuan2017temporal}: train on 7,986 videos (54,000 clips) and evaluate on 5,000 randomly selected test frames from 200 test videos.

The EPIC-KITCHENS \cite{Damen2018EPICKITCHENS} is an ego-centric video dataset which is captured by head-mounted camera in different kitchen scenes. In the training data, the action class is annotated for 28,473 trimmed video clips and the object bounding boxes are labeled for 331 object classes. As the object bounding box annotations are not provided for the test splits, we divide the training data into training, validation and test parts. The top 15 frequent object classes (which are present in 85 action classes) are selected for experiments, resulting in 8,520 training, 1,000 validation and 200 test video clips. We randomly sample three times from each training clip and generate 28,560 training samples. We also randomly sample 1,200 test frames from the test clips.


\textbf{Image dataset} The HICO-DET dataset \cite{chao2018learning} is designed for human-object interaction (HOI) detection task. This dataset includes 38,118 training images and 9,658 test images. The human bounding box, object bounding box and an HOI class label are annotated for both training and test images. In total, there are 80 object classes (\eg~\textit{cup}, \textit{dog}, \etc) and 600 HOI classes (\eg~\textit{hold cup}, \textit{feed dog}, \etc).  We filter out all samples with ``no\_interaction'' HOI labels, interaction class with less than 20 training samples and all ``person'' as object class samples. This results in 32,100 training samples of 510 interaction classes and 79 object classes. We use the HOI labels as action class labels during training and the object bounding box annotations are used only for evaluation. Unlike Charades where the interactions mostly happen between one subject and one object, there are cases where multiple people interact with one object (\eg~``boarding the airplane'') and one person interacts with multiple objects (\eg~``herding cows''), which makes it more challenging to learn the object appearance.

We report per-class \textit{average precision} (AP) at \textit{intersection-over-union} (IoU) of 0.5 between detection and ground truth boxes, and also mean AP (mAP) as a combined metric, following the tradition of \cite{yuan2017temporal}. We also report \textit{CorLoc} \cite{deselaers2012weakly}, a commonly-used weakly supervised detection metric. CorLoc represents the percentage of images where at least one instance of the target object class is correctly detected (IoU\textgreater0.5) over all images that contain at least one instance of the class.

\vspace{-0.4\baselineskip}
\subsection{Implementation details}
\vspace{-0.4\baselineskip}
We use VGG-16 and ResNet-101 pre-trained on ImageNet dataset as our backbone feature extraction networks. All \textit{conv} layers in the network are followed with \textit{ReLU} activation except for the top classification layer. Batch normalization \cite{ioffe2015batch} is applied after all convolutional layers. In order to compute the classification scores ($s_O, s_A^H, s_A^O$), three branches are built on top of the last convolutional block. Each branch consists of ROI-pooling layer and 2-layer multiple layer perception (MLP) of intermediate dimension of 4096. The threshold for removing person proposal regions is set as $\theta_h=0.5$. Selective search \cite{uijlings2013selective} is used to extract object proposals for all our experiments.


The Adam optimizer \cite{kingma2014adam} is applied with learning rate of $2\times 10^{-5}$ and batch size of $4$. The loss weights are set as $\alpha_a=1.0$, $\alpha_o=2.0$.
The number of sampled frames in a clip is set as $n=8$ and the number of proposals is set as $n_r=700$. The whole framework is implemented with PyTorch \cite{paszke2017automatic}. We train on a single Nvidia Tesla M40 GPU. The whole training converges in 20 hrs. More details of implementation are presented in supplemental material.

\vspace{-0.3\baselineskip}
\subsection{Influence of modeling spatial location of object}
\vspace{-0.4\baselineskip}
\label{sec:ablation}

Unlike many existing methods for weakly supervised object detection, our framework explicityly models the spatial locaition of the object \wrt to the detected person and encodes it into two different loss functions in Eq.~\ref{eqn:obj_cls_loss},~\ref{eqn:act_cls_loss}. We explore the effect of modeling this spatial prior through different distributions and its contribution to each of the loss terms.

The different distributions include:  (a) normal distribution, (b) a fixed grid of probability values, where we make a discrete version of spatial prior module by pre-defining a $3\times 3$ grid around the keypoint, and (c) a simple center prior where we penalize object detections farther away from the center of the object. Note that, we totally removed person detection bounding box and pose estimation in the center prior baseline. For this baseline, we use the frame center as the anchor location $\scr{L}$ and learn the $\mu_a$ and $\sigma_a$.

We also experimented with learning distribution mean only ($\mu_a$), learning variance only ($\sigma_a$) and joint learning of mean and variance ($\mu_a+\sigma$) for the normal distribution. We alos experimented with using only object classification or action classification loss.

\begin{table}[]
\vspace{-0.8\baselineskip}
\fontsize{8}{9}\selectfont
\setlength{\tabcolsep}{3pt}
\centering
\caption{Detection performance of different variants on Charades}
\label{tbl:ab_study}
\begin{tabular}{l|c|c|c}
\specialrule{.2em}{.1em}{.1em}
 Spatial prior  & Loss term  & mAP  & CorLoc   \\ \hline
 Center            & action+object  & 3.43 & 34.27\\
 Grid              & action+object  & 4.32 &  36.94\\ \hline
 Normal ($\mu$)     & action+object  & 6.27  & 42.36\\
 Normal ($\sigma$)    & action+object  & 4.86 & 38.05\\
 Normal ($\mu + \sigma$)  & action  & 2.61 &  31.60\\
 Normal ($\mu + \sigma$)    & object  & 5.86 &  39.24 \\
 Normal ($\mu + \sigma$)    & action+object  & \textbf{8.76} & \textbf{47.91}\\ \hline
\end{tabular}
\vspace{-0.8\baselineskip}
\end{table}

The quantitative results of VGG-16 as the backbone network are presented in \tabref{tbl:ab_study} for different ablation settings. First, we observe that a learnable grid-based or normal distribution for the anchor location outperforms a simple heuristic choice of the image center as the anchor. We also see that the normal distribution, where both mean and variance are learned for each action-object pair leads to better results compared to the other settings. This shows that good modeling of the object spatial prior \wrt~human in an action provides strong cues for detection. We also notice that jointly modelling both action and object classification achieves the best result.

\begin{figure}
\centering
\vspace{-0.4\baselineskip}
\includegraphics[width=0.45\textwidth]{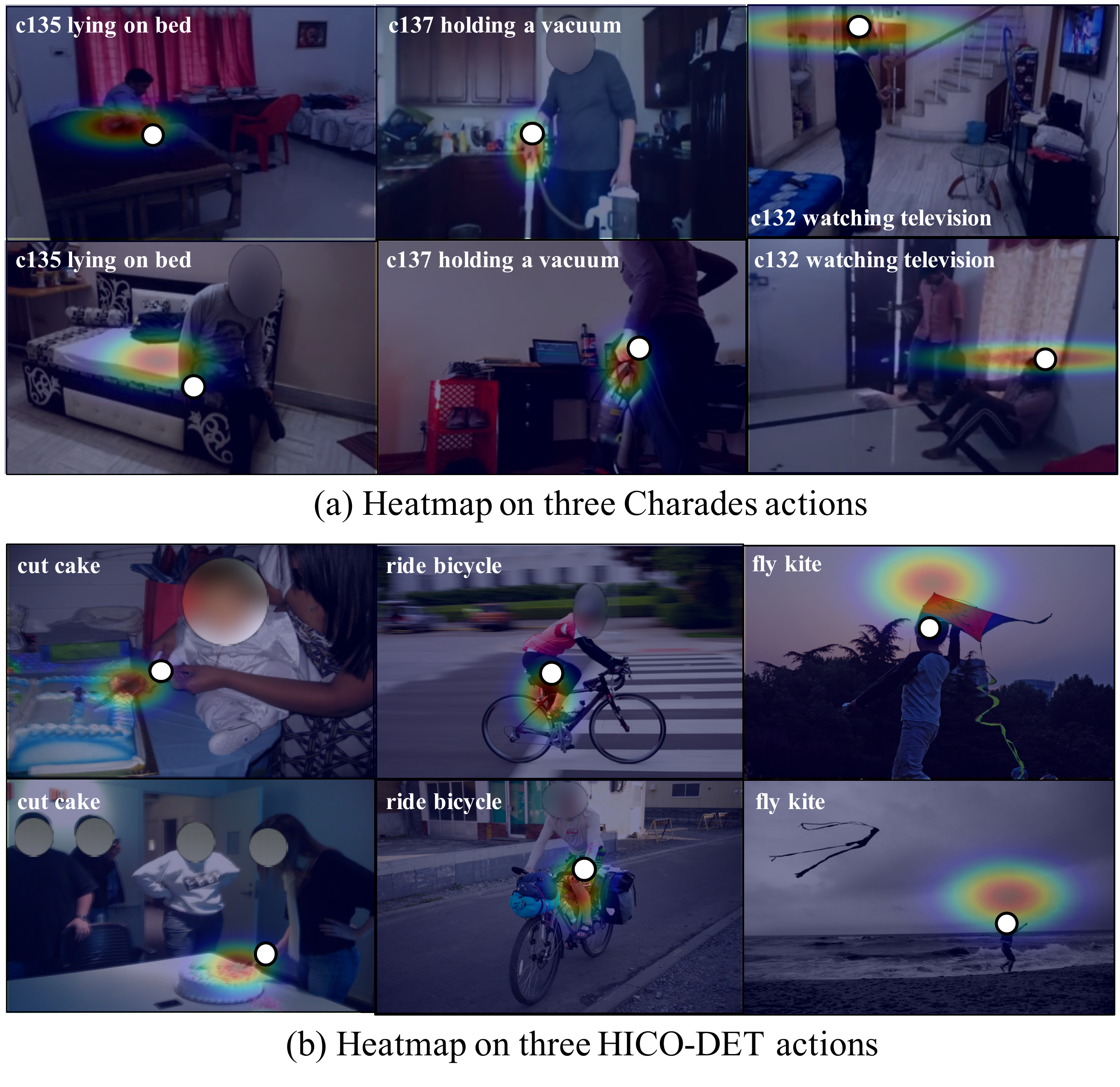}
\caption{Visualization of learned object location probability w.r.t. selected person keypoint. The heatmap represents the probability of object location (brighter color represents larger probability value) and the white circle represents the selected keypoint.}
\label{fig:distribution}
\vspace{-0.3\baselineskip}
\end{figure}

We also visualize the learned distribution of object location probabilities from the prior module for a few sample videos/images in Fig. \ref{fig:distribution}. The learned distribution often has large probability weights around the object mentioned in the action. For example, in the first two columns of the visualization, it is much easier to localize the object with the cues from the heatmap. However, we also note that this distribution is less useful for actions where there is no consistent physical interaction between the human and the object. This is shown in the last column of the figure, for actions like ``watching television" and ``flying kite". Our approach reports relatively low mAP performance on such object classes (Tab. \ref{tbl:class_wise_charades} and Tab. \ref{tbl:class_wise_hico}).

\vspace{-0.6\baselineskip}
\subsection{Comparison with existing methods}
\vspace{-0.4\baselineskip}
We compare our method with other weakly supervised methods and their variants: (1) WSDDN \cite{bilen2016weakly}; (2) ContextLocNet \cite{kantorov2016contextlocnet}; (3) PCL \cite{tang2018pcl}; action-driven weakly supervised object detection method: (4) TD-LSTM \cite{yuan2017temporal} and (5) R*CNN \cite{gkioxari2015contextual} which is designed for action recognition with awareness of the main object. We used the main object bounding box as the object detection result. R*CNN is pre-trained on Pascal-action dataset and then finetuned on Charades or HICO-DET dataset. Note that existing methods (1), (2), (3), (4) do not use person bounding box or keypoint detection results in their model unlike our method. While (5) uses person bounding box, it doesn't use person keypoints.

The person detection and pose models used in our model were trained only once and kept fixed during training. The annotations required to train the person models are very inexpensive in comparison to fully supervised models which need bounding box annotation for every object class. The resource demands of annotating person bounding boxes and pose is amortized across all object classes. However, since these models are not used in traditional weakly supervised methods, we enable fair comparison by constructing variants of PCL and R*CNN: (6) R*CNN with spatial prior and (7) PCL with spatial prior, where we replace the max pooling in R*CNN and mean pooling in PCL with a weighted sum where the weights are computed from spatial prior distribution as in our implementation (more details are presented in supplemental material).

\begin{table*}[t]
\centering
\fontsize{8}{8.5}\selectfont
\caption{AP performance (\%) on each object class and mAP (\%) comparison with different weakly supervised methods on Charades.}
\label{tbl:class_wise_charades}
\def\arraystretch{1.1}
\setlength{\tabcolsep}{2.5pt}
\begin{tabular}{l|ccccccccccccccccc|c}
\specialrule{.2em}{.1em}{.1em}
Methods                                        & bed & broom & chair & cup & dish & door & laptop & mirror & pillow & refri & shelf & sofa    & table   & tv   & towel       & vacuum    & window     & mAP(\%)      \\ \hline
WSDDN \cite{bilen2016weakly}                   & 2.38 & 0.04 &1.17 &0.03 & 0.13 & 0.31 & 2.81 & 0.28 & 0.02 & 0.12 & 0.03 & 0.41 & 1.74 & 1.18 & 0.07 & 0.08 & 0.22 & 0.65   \\
R*CNN \cite{gkioxari2015contextual}           & 2.17 & 0.44 & 2.03 & 0.31 & 0.08 & 0.77 & 2.64 & 0.32 & 1.24 & 2.36 & 0.82 & 1.41 & 0.65 & 0.72 & 0.07 & 0.65 & 0.17 & 0.99 \\
ContextLocNet \cite{kantorov2016contextlocnet} & 7.40 & 0.03 & 0.55 & 0.02 & 0.01 & 0.17 & 1.11 &0.66 & 0.00 & 0.07 & 1.75 & 4.12 & 0.63 & 0.99 & 0.03 & 0.75 & 0.78 & 1.12  \\
TD-LSTM \cite{yuan2017temporal}                & 9.19 & 0.04 & 4.18 & 0.49 & 0.11 & 1.17 & 2.91 & 0.30 & 0.08 & 0.29 & 3.21 & 5.86 & 3.35 & 1.27 & 0.09 & 0.60 & 0.47 & 1.98 \\
PCL \cite{tang2018pcl}                         & 4.62 & 1.07 & 2.21 & 1.26 & 1.08 & 2.49 & 3.61 & \textbf{5.13} & 1.34 & 4.46 & 3.29 & 5.61 & 3.84 & 3.26 & 1.17 & 1.43 & 2.27 & 2.83 \\
R*CNN + prior                            & 6.82 & 3.64 & 5.39 & 3.25 & 2.47 & 3.36 & 5.27 & 1.07 & 2.38 & 6.34 & 3.29 & 5.72 & 4.09 & 1.03 & 1.26 & 3.41 & 0.86 & 3.50 \\
PCL + prior                                    & 10.57 & 5.63 & 8.24 & 3.52 & 3.71 & 5.63 & 6.86 & 4.96 & 5.23 & 11.39 & 4.88 & 10.46 & 6.32 & \textbf{3.53} & 4.06 & 4.89 & \textbf{3.07} & 6.05 \\
\hline
Ours-vgg-16 (w/o prior)                         & 6.71 & 2.32 & 5.48 & 2.49 & 1.04 & 3.60 & 4.02 & 3.42 & 4.39 & 7.76 & 3.15 & 7.43 & 3.26 & 1.62 & 0.89 & 2.24 & 1.23 & 3.60\\
Ours-vgg-16                                  & 14.92 & 10.23 & 13.08 & 7.65 & 5.21 & 6.44 & 8.65 & 4.79 & 9.14 & 18.07 & 7.29 & 17.21 & 8.46 & 2.37 & 5.46 & 7.23 & 2.64 & 8.76\\
Ours-ResNet-101                            & \textbf{16.54} & \textbf{11.63} & \textbf{14.87} & \textbf{8.62} & \textbf{6.73} & \textbf{8.29} & \textbf{11.32} & \textbf{4.96} & \textbf{9.81} & \textbf{19.24} & \textbf{9.03} & \textbf{18.49} & \textbf{9.86} & 3.05 & \textbf{6.48} & \textbf{8.08} & 3.02 & \textbf{10.03}\\\hline
\end{tabular}
\vspace{-0.4\baselineskip}
\end{table*}

\begin{figure*}
\vspace{-0.2\baselineskip}
\centering
\includegraphics[width=0.95\textwidth]{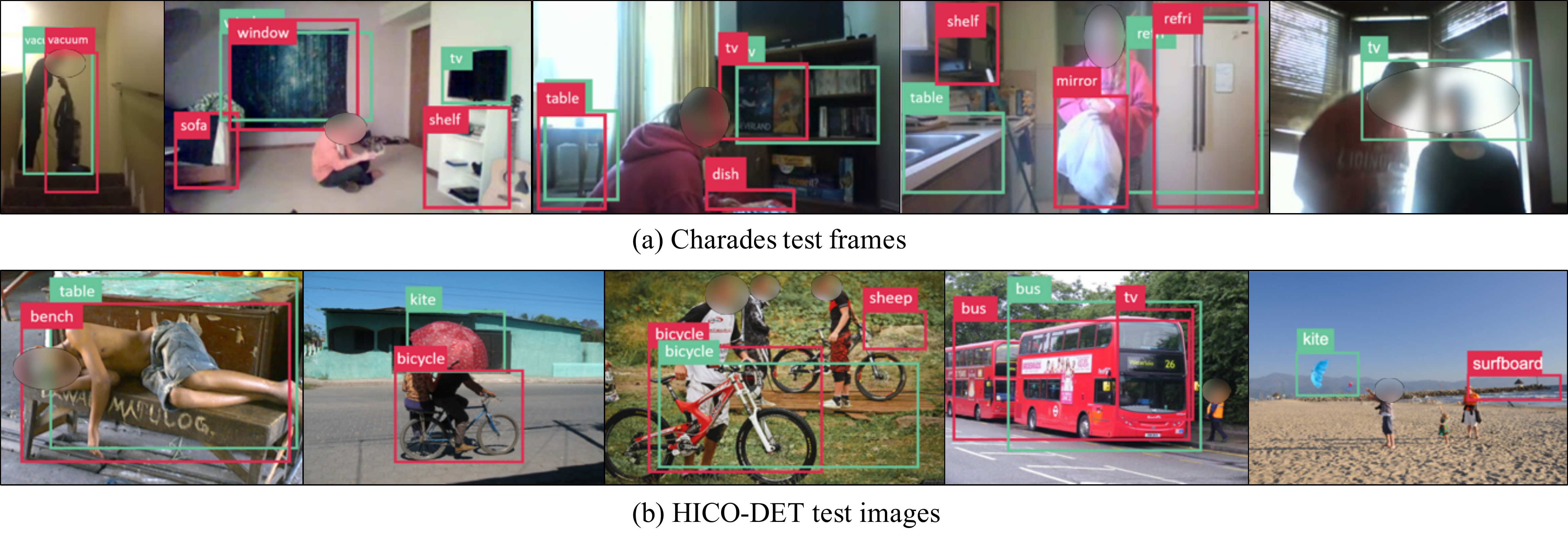}
\caption{Qualitative detection results on (a) Charades test frames and (b) HICO-DET test images. Red bounding boxes denote our results and green bounding boxes denote results of PCL \cite{tang2018pcl}}
\label{fig:det_vis}
\vspace{-0.8\baselineskip}
\end{figure*}

Results from TD-LSTM \cite{yuan2017temporal} are shown only for Charades, since it is a video-specific model and code is not available. Also, we report results from weakly-supervised models whose code is available or whose results on Charades, HICO-DET or EPIC KITCHENS datasets is readily available. Also, note that many methods such as \cite{diba2017weakly,wei2018ts2c,bai2017multiple} are built on top of the vanilla WSDDN method by adding signals such as segmentation, contextual information, instance refinement, \etc and these ideas are complementary to the ones presented in this work, and can be added to our model to achiever better results.
The per-class AP and combined mAP performances on the two datasets are presented in Tab. \ref{tbl:class_wise_charades} and Tab. \ref{tbl:class_wise_hico} respectively. 10 object classes on HICO-DET are randomly selected and presented.

\begin{table*}[]
\vspace{-0.5\baselineskip}
\centering
\fontsize{8.5}{9}\selectfont
\caption{AP performance (\%) on selected object classes and mAP (\%) comparison with other weakly supervised methods on HICO-DET.}
\label{tbl:class_wise_hico}
\def\arraystretch{1.1}
\setlength{\tabcolsep}{3pt}
\begin{tabular}{l|cccccccccc|c}
\specialrule{.2em}{.1em}{.1em}
Methods & apple & bicycle & bottle & chair & cellphone & frisbee & kite & surfboard & train & umbrella & mAP(\%)      \\ \hline
R*CNN \cite{gkioxari2015contextual}   & 1.13 & 3.26 & 1.57 & 2.35 & 1.47 & 1.02& 0.32 & 2.70 & 2.86 & 3.04 & 2.15 \\
WSDDN \cite{bilen2016weakly}  & 1.46 & 5.19 & 1.52 & 3.87 & 2.02 & 2.44 & 1.15 & 2.86 & 6.76 & 3.35 & 3.27   \\
PCL \cite{tang2018pcl}          & 1.27 & 5.82 & 2.31 & 2.84 & 3.06 & \textbf{3.11}& 1.16 & 2.60 & 7.93 & 3.47 & 3.62 \\
PCL + prior                 & 2.06 & 6.49 & 2.54 & 3.69 & 5.14 & 2.96 & \textbf{1.37} & 4.06 & 8.13 & 4.87 & 4.19 \\
\hline
Ours-vgg-16 (w/o prior)                    & 1.23 & 5.15 & 1.19 & 3.47 & 3.82 & 2.24 & 0.73 & 3.65 & 6.22 & 3.14 & 3.16\\
Ours-vgg-16                       & \textbf{2.47} & \textbf{8.64} & \textbf{3.59} & \textbf{5.74} & \textbf{7.36} & 2.85 & 0.87 & \textbf{7.29} & \textbf{8.47} & \textbf{6.63} & \textbf{5.39}\\ \hline
\end{tabular}
\vspace{-1.2\baselineskip}
\end{table*}

On Charades dataset, our method achieves 6\% mAP boost compared to PCL \cite{tang2018pcl}. Our method performs better on object classes like broom, refrigerator, vacuum, \etc. The spatial prior patterns of the interactions involving these object classes are more predictable and thus the prior modeling benefits our approach more than on other object classes. For object like tv, the spatial prior pattern of the interaction (\eg \textit{watch tv}) is more diverse and thus difficult to model, resulting only a small boost in mAP. The same performance pattern also applies to HICO-DET dataset. On the object class kite, our method slightly performs inferior to the baseline method.

\begin{table}[]
\vspace{-0.2\baselineskip}
\centering
\fontsize{8.5}{9}\selectfont
\caption{mAP (\%) comparison with other weakly supervised methods on EPIC KITCHENS}
\label{tbl:epic}
\def\arraystretch{1.2}
\begin{tabular}{l|cc}
\specialrule{.2em}{.1em}{.1em}
Methods                        & mAP  & CorLoc\\ \cline{1-3}
R*CNN \cite{gkioxari2015contextual} & 2.54 & 32.68  \\
PCL \cite{tang2018pcl}   & 4.68 & 40.64 \\
PCL + prior  & 6.82 & 46.69  \\ \cline{1-3}
Ours-vgg-16 & \textbf{9.75} & \textbf{52.53}  \\ \cline{1-3}
\end{tabular}
\vspace{-1.8\baselineskip}
\end{table}

We observe that the spatial prior from our model is effective in localizing the object during training even when combined with other models such as R*CNN \cite{gkioxari2015contextual} and PCL \cite{tang2018pcl}. R*CNN with spatial prior modeling outperforms TD-LSTM, which is specifically designed for the action driven weakly supervised object detection task.

We also report our model's performance without the spatial prior module (ours (w/o prior)). This variant of the model doesn't require any person bounding box and keypoint information, and is directly comparable to existing weakly supervised methods. We note that even without these signals, our model can outperform existing methods. This can be attributed to the fact that our model jointly uses both action and object labels during training. It identifies the object bounding box which can both help action classification and object classification during training.

The qualitative comparison between our method and PCL is presented in Fig. \ref{fig:det_vis}. Our approach localizes the object more accurately. Multiple object classes and multiple instances can be detected through our trained object classification stream. The last column shows our failure cases. On Charades, both PCL \cite{tang2018pcl} and our method fails to detect the windows and on HICO-DET, our method fails to localize the kite. One possible reason is that actions like ``watch out of the window" do not have direct human-object interaction.

Our approach is also extended to ego-centric EPIC KITCHENS datset. Since human keypoints are not visible in this dataset, we applied ``center'' spatial prior modeling used in Sec.~\ref{sec:ablation}. As the camera is fixed with respect to the human, the anchor location is already implicitly modeled by this center prior. We compare with R*CNN \cite{gkioxari2015contextual} and PCL \cite{tang2018pcl} on the 1,200 test frames. Egocentric videos have a strong prior for object spatial locations and hence our method is able to outperform other methods in Tab. \ref{tbl:epic}.

\vspace{-0.5\baselineskip}
\subsection{Effect of supervision in training}
\vspace{-0.4\baselineskip}
\label{sec:supervision}

Weakly supervised object detection aims to train object detection models without any bounding box labels. However, in practice it is easy and efficient to annotate at least a few bounding boxes in training images/videos. This is similar to low-shot and semi-sueprvised settings. We believe that it is important to test weakly supervised approaches in such a practical setting as well.

To this end, we explore the effect of adding varying amounts of ground truth object bounding box annotations into our training data. We achieve this by augmenting the losses described in Sec. \ref{sec:approach}, with an additional supervised object detection loss for videos/images where bounding box annotations are available. This loss is the same as the traditional object detection loss used in Fast-RCNN.

In practice, the IoU between object proposals and ground truth object bounding boxes is calculated and proposals having higher IoU than the threshold are considered positive samples and rest as negative. The threshold IoU is set as 0.45 to guarantee a reasonable positive samples per image. The negative and positive sample ratio is set as 5.

We compare with two baselines: (1) model without weak supervision: model trained only with supervised detection loss on images/videos with bounding box annotations and without any weakly supervised data (Ours (w/ only strong supervision)),  and (2) R*CNN \cite{gkioxari2015contextual} with additional object bounding box supervision as above (R*CNN (w/ strong+weak supervision)).

\begin{figure}
\centering
\includegraphics[width=0.5\textwidth]{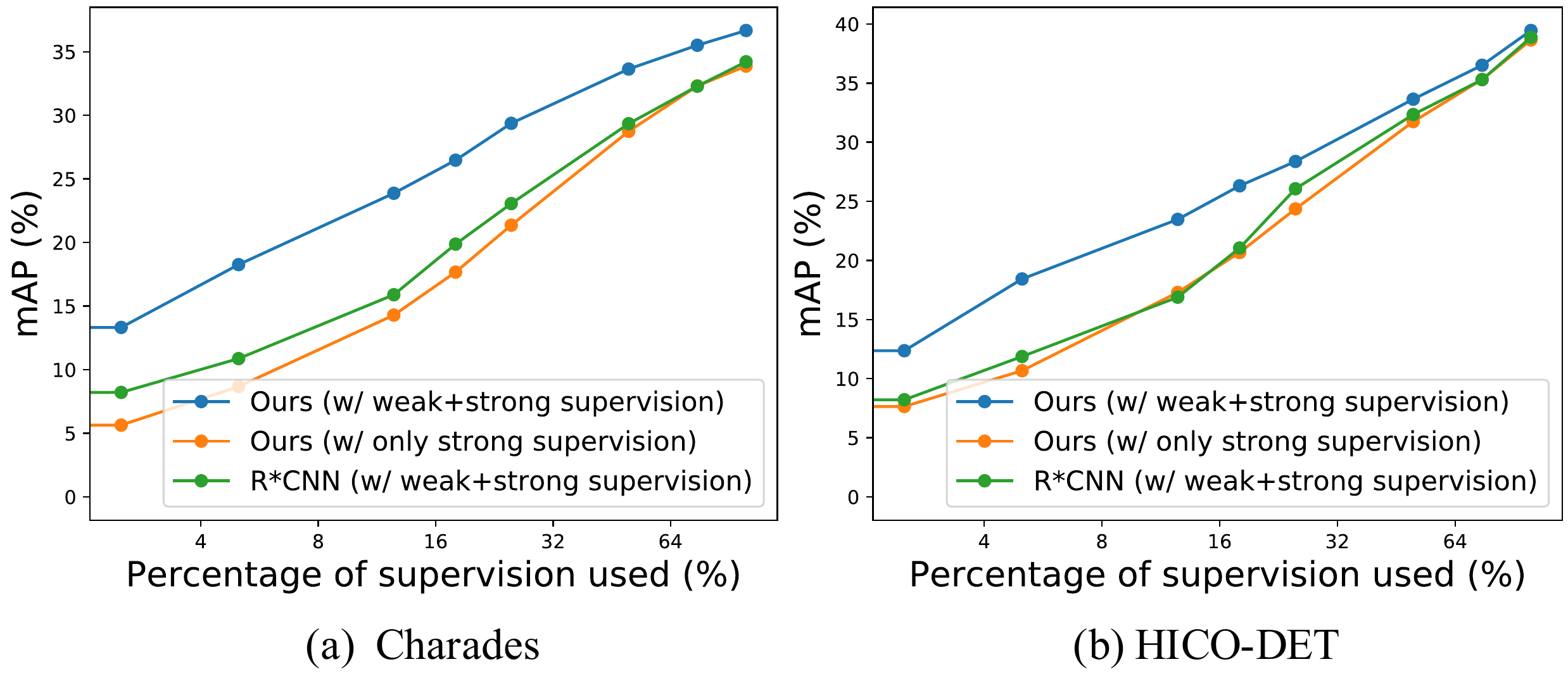}
\caption{Performance comparison between our method trained with different supervision settings and R*CNN trained with both strong and weak supervision on (a) Charades and (b) HICO-DET.}
\label{fig:semi-supervision}
\vspace{-1.3\baselineskip}
\end{figure}

We evaluate this setup on both Charades and HICO-DET datasets. The quantitative results are presented in Fig. \ref{fig:semi-supervision}. The x-axis (log scale) represents the percentage of training samples with object bounding box annotations. For example, the point $x\%$ represents that for a random $x\%$ of training data samples, the bounding box annotations are present. The remaining training samples, only have action class label. Note that $0\%$ is the weakly-supervised setting considered earlier, while $100\%$ represents fully-supervised setting. We observe that the mAP increases log-linearly as more supervision is added to the training.


For Charades, when small amount of supervision is added, we observe that our model which uses additional weakly-supervised data outperforms the model without any weak supervision. This clearly shows the potential of our weakly-supervised approach to provide complementary value in a low-shot detection setting. With as low as 70\% supervision, our approach already matches the performance of fully-supervised method at 100\% supervision. This means that we could cut down the amount of supervision needed to train the model without sacrificing performance. As expected, the gap between the two approaches decrease with increase in supervision. Even with ``100\%'' bounding box annotations, our model still outperforms the fully supervised method by $2$ mAP points due to joint training with action and object classification loses.

We also observe that performance gap is smaller for images (HICO-DET). We believe weak supervision is more effective in videos compared to images, where temporal linking of proposals helps in avoiding spurious detections during training.

\vspace{-0.6\baselineskip}
\section{Conclusion}
\vspace{-0.5\baselineskip}
We observe that object spatial location, appearance and movement are tightly related to the action performed with the object in images and videos. We propose a model that leverages these observations to train object detection models from samples annotated only with action labels. Comprehensive experiments are conducted on both video and image datasets. The comparison with SoTA methods shows that out approach outperforms existing weakly supervised approaches. Further our approach can also help reduce the amount of supervision required for object detection models.


\end{document}